\documentclass[conference]{IEEEtran}
\IEEEoverridecommandlockouts
\AtBeginDocument{\usepackage{epstopdf}}
\usepackage[utf8]{inputenc}
\usepackage{cite}
\usepackage{amsmath,amsfonts}
\usepackage{algorithmic}

\usepackage{graphicx}
\usepackage{subcaption}
\usepackage{textcomp}
\usepackage{xcolor}
\usepackage{booktabs}  
\usepackage{adjustbox} 
\usepackage{multirow}
\usepackage{caption}
\usepackage{makecell}
\usepackage{newtxtext}
\usepackage{newtxmath}
\def\BibTeX{{\rm B\kern-.05em{\sc i\kern-.025em b}\kern-.08em
    T\kern-.1667em\lower.7ex\hbox{E}\kern-.125emX}}
\usepackage[colorlinks=true, linkcolor=blue, urlcolor=blue, citecolor=blue]{hyperref}

\begin{document}

\title{IntersectioNDE: Learning Complex Urban Traffic Dynamics based on Interaction Decoupling Strategy }

\author{
Enli Lin$^{1}$\thanks{$^{1}$Department of Mechanical Engineering, Tsinghua University, Beijing, China. 
Email: lel22@mails.tsinghua.edu.cn}, 
Ziyuan Yang$^{2}$\thanks{$^{2}$Department of Automation, Tsinghua University, Beijing, China. 
Email: yangziyu22@mails.tsinghua.edu.cn, qiujinglu@mail.tsinghua.edu.cn,  hujm@mail.tsinghua.edu.cn,
fshuo@tsinghua.edu.cn}, 
Qiujing Lu$^{2}$,
Jianming Hu$^{2}$,
Shuo Feng$^{2*}$\thanks{*Corresponding author}}

\maketitle
\begin{abstract}
Realistic traffic simulation is critical for ensuring the safety and reliability of autonomous vehicles (AVs), especially in complex and diverse urban traffic environments. 
However, existing data-driven simulators face two key challenges:  a limited focus on modeling dense, heterogeneous interactions at urban intersections---which are prevalent, crucial, and practically significant in countries like China, featuring diverse agents including motorized vehicles (MVs), non-motorized vehicles (NMVs), and pedestrians---and the inherent difficulty in robustly learning high-dimensional joint distributions for such high-density scenes, often leading to mode collapse and long-term simulation instability.
We introduce City Crossings Dataset (CiCross), a large-scale dataset collected from a real-world urban intersection, uniquely capturing dense, heterogeneous multi-agent interactions, particularly with a substantial proportion of MVs, NMVs and pedestrians.
Based on this dataset, we propose IntersectioNDE (Intersection Naturalistic Driving Environment), a data-driven simulator tailored for complex urban intersection scenarios. 
Its core component is the Interaction Decoupling Strategy (IDS), a training paradigm that learns compositional dynamics from agent subsets, enabling the marginal-to-joint simulation. 
Integrated into a scene-aware Transformer network with specialized training techniques, IDS significantly enhances simulation robustness and long-term stability for modeling heterogeneous interactions.
Experiments on CiCross show that IntersectioNDE outperforms baseline methods in simulation fidelity, stability, and its ability to replicate complex, distribution-level urban traffic dynamics.

\end{abstract}

\section{Introduction}
Ensuring the safety and reliability of autonomous vehicles (AVs) remains a critical challenge, requiring extensive validation across diverse, complex driving scenarios~\cite{feng2023dense}. Simulation-based testing is widely adopted for its scalability, cost efficiency, and ability to explore safety-critical edge cases that are impractical or unsafe to encounter in reality~\cite{feng2021intelligent, li2019aads}.

Recent advances in data-driven traffic simulation have shown promise by learning behavior directly from real-world data. However, existing methods exhibit a limited focus on modeling complex urban environments at busy intersections, especially those in countries like China featuring dense, heterogeneous traffic with motorized vehicles (MVs), non-motorized vehicles (NMVs), and pedestrians. 
Furthermore, a core challenge persists in the inherent difficulty of robustly learning high-dimensional joint distributions of full-scene states from limited data, which often leads to mode collapse and instability in long-term closed-loop simulations~\cite{guo2024lasil}. This is further exacerbated by representing diverse interaction patterns across heterogeneous agent types within a single unified model.

To address these challenges, we present two key contributions. First, we introduce City Crossings Dataset (CiCross),a large-scale real-world urban intersection dataset capturing dense, heterogeneous interactions. Second, building on CiCross, we propose IntersectioNDE (Intersection Naturalistic Driving Environment), a data-driven scene-level simulator for robust, stable urban traffic. Its core, the Interaction Decoupling Strategy (IDS), partitions the full scene into disjoint agent subsets based on interaction criteria for compositional training via sampling, while implicitly synthesizing full future scenes autoregressively during inference. This compositional learning paradigm enables more effective generalization to unseen interaction combinations within the same scene and alleviates the need to directly model full-scene joint distributions. IntersectioNDE is implemented via a scene-aware interaction Transformer network, coupled with specialized training techniques that capture complex interactions and maintain long-term stability.

Experiments on CiCross show that IntersectioNDE exhibits strong heterogeneous interaction modeling capabilities and simulation robustness, as well as its potential for replicating the distribution-level characteristics of real-world urban traffic.

\begin{figure*}[htbp]
    \centering
    \includegraphics[width=18cm, trim=2cm 5.5cm 2cm 5cm, clip]{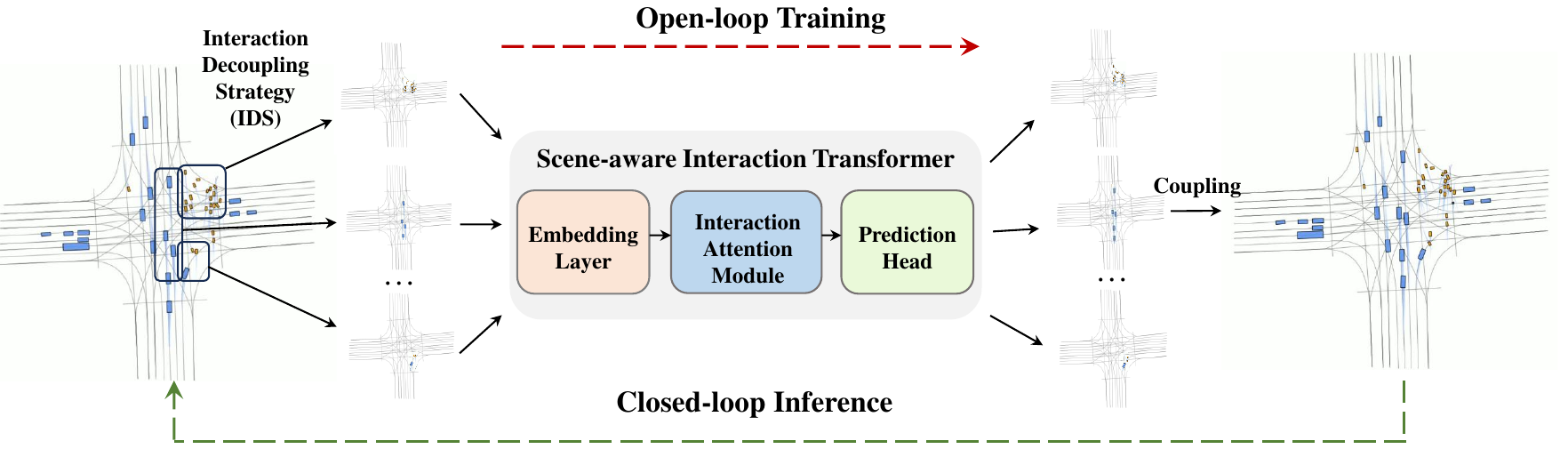}
    \caption{Overview of the training and inference pipeline of IntersectioNDE.}
    \label{fig:pipeline}
\end{figure*}

\section{RELATED WORK}
\subsection{Traffic Datasets for Simulation}
High-quality, large-scale traffic datasets are indispensable for developing and evaluating data-driven autonomous driving systems, particularly for perception, prediction, and simulation tasks. 
Recent years have witnessed the release of numerous benchmark datasets, offering diverse driving scenarios. 
Prominent examples include the Waymo Open Dataset~\cite{waymo2021large}, nuScenes~\cite{caesar2020nuscenes}, Argoverse~\cite{chang2019argoverse}, INTERACTION~\cite{zhan2019interaction}, NGSIM~\cite{kovvali2007video}, and Lyft Level 5 Perception Dataset~\cite{houston2021lyft}, containing traffic data across multiple scenarios in urban and suburban environments.
Datasets such as InD~\cite{bock2020ind}, HighD~\cite{krajewski2018highd}, and RounD~\cite{krajewski2020round} provide traffic data for single specific scenarios, such as intersections, highways, and roundabouts. However, despite their scale and value, these datasets often contain a limited variety of traffic participants and lack sufficient representation of dense interactions in complex urban intersections, particularly those involving MVs, NMVs and pedestrians. This constrains the heterogeneity and diversity of traffic simulation behaviors modeled from such data.
\subsection{Traffic Simulation}
Traffic simulation methods can be broadly categorized into rule-based and data-driven approaches. Rule-based simulators, such as SUMO~\cite{sumo2018microscopic}, VISSIM~\cite{vissim2010microscopic} and MITSIM~\cite{yang1996microscopic} rely on predefined physical rules and behavioral models. 
While useful for macroscopic traffic flow analysis, their reliance on fixed parameters often limits their ability to reproduce the diversity and complexity of real-world, naturalistic driving behaviors and heterogeneous interactions at a microscopic level.
Consequently, data-driven traffic simulation methods, which aim to learn traffic dynamics directly from observed data, have become increasingly prevalent. 
Within data-driven simulation, two main paradigms exist: agent-centric and scene-level. 

Agent-centric approaches~\cite{nayakanti2023wayformer, shi2022motion, shi2024mtr++, wang2023multiverse} typically learn individual agent behaviors and iteratively update agent states within the simulated environment. While conceptually straightforward, this paradigm often suffers from low simulation efficiency due to its sequential and iterative nature~\cite{ye2022efficient}, and faces challenges in coordinating complex interactions among numerous agents.

In contrast, scene-level data-driven methods operate as generative models that learn to directly output the next state of the entire traffic scene given its current state and history. These approaches aim to capture the holistic scene dynamics and generate future states auto-regressively. 
From the perspective of generative model architecture, this includes deep Autoregressive Models~\cite{feng2023trafficgen86, tan2021scenegen87}, Variational Autoencoders (VAE)~\cite{rempe2022generating89, tang2021exploring91}, Generative Adversarial Networks (GAN)~\cite{bergamini2021simnet67, zhang2022systematic94}, and Diffusion Models~\cite{zhong2023guided96, pronovost2023generating97, wang2024drivedreamer}.
A notable example is NeuralNDE~\cite{yan2023learning}, which successfully demonstrates statistical realism in specific scenarios like roundabouts. 

However, extending such models to the complex urban intersections prevalent in countries like China, with their dense, highly unstructured interactions among motorized and non-motorized vehicles, still warrants further research. Moreover, scene-level models face challenges in maintaining robustness and long-term stability during simulation, mainly due to the high-dimensional nature of dense traffic dynamics which leads to data sparsity and potential model drift or collapse~\cite{chen2024data}.

\begin{figure*}[htbp]
    \centering
    \includegraphics[width=18cm, trim=1cm 5.5cm 1cm 5cm, clip]{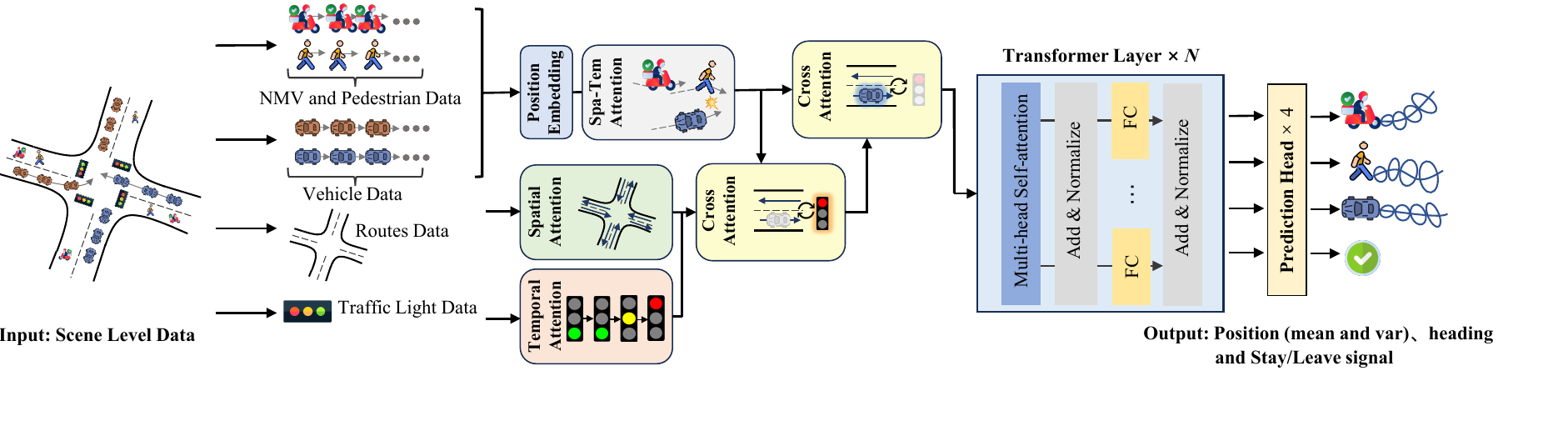}
    \caption{Architecture of the Scene-Aware Interaction Transformer Model.}
    \label{fig:network}
\end{figure*}

\section{METHOD}
\subsection{Problem Formulation}
\label{sec:method:problem_formulation}

The task of complex urban intersection traffic simulation is framed as learning a generative model capable of producing realistic future scene states over a prediction horizon $T_{pred}$. Let $A_{\tau} = \{a_1, \dots, a_{N_{\tau}}\}$ be the set of $N_{\tau}$ agents present at time $\tau$. The state of an agent $a_j$ at time $\tau$ is $s_{j, \tau} \in \mathcal{S}_{\text{agent}}$, where $\mathcal{S}_{\text{agent}}$ defines the agent's state space (e.g., kinematics, category). The agent states in a scene at time $\tau$ are denoted by $S_{\tau} = \{s_{j, \tau} \mid a_j \in A_{\tau}\}$. A complete scene instance at time $\tau$, denoted by $G_{\tau}$, comprises the agent states $S_{\tau}$, the static map information $M \in \mathcal{M}$, and the dynamic traffic light state $L_{\tau} \in \mathcal{L}$. We will represent a scene instance as a tuple $(S_{\tau}; M, L_{\tau})$, explicitly separating the variable agent states from the context. We are given a sequence of historical scene instances $G_{t-T_{hist}+1:t} = \{(S_{\tau}; M, L_{\tau}) \mid \tau \in [t-T_{hist}+1, t]\}$.

The standard formulation for this task is to learn the conditional probability distribution of the subsequent scene instances over the prediction horizon $G_{t+1:t+T_{pred}}$ given the historical inputs: $P_{\text{data}}(G_{t+1:t+T_{pred}} \mid G_{t-T_{hist}+1:t})$. Training a model $F_{\theta}$ to approximate this full joint distribution over the entire scene state space for the future horizon aims to enable realistic closed-loop simulation. However, directly learning this high-dimensional joint distribution from finite data poses significant challenges regarding model robustness and generalization to novel, complex scenes. To address these challenges, we propose a novel modeling approach.

\subsection{ Interaction Decoupling Strategy (IDS)}
\label{sec:method:ids} 

To address the challenges associated with directly learning the high-dimensional joint scene distribution over a future horizon from finite data and to enhance model robustness, we propose the Interaction Decoupling Strategy (IDS). 
IDS enables a marginal-to-joint simulation capability by learning the dynamics of various interacting agent subsets.

\subsubsection{IDS Training}
\label{sec:method:ids:training}

Under the IDS for training, for a sequence of true scene instances $G_{t-T_{hist}+1:t+T_{pred}}^{\text{GT}} = \{(S_{\tau}^{\text{GT}}; M^{\text{GT}}, L_{\tau}^{\text{GT}}) \mid \tau \in [t-T_{hist}+1, t+T_{pred}]\}$ from the data distribution $\mathcal{D}_{\text{data}}$, we consider the scene instance at time $t$, $G_t = (S_t; M, L_t)$. We partition the set of agents $A_t$ at time $t$ into $k$ disjoint interaction groups based on predefined spatial and behavioral criteria (e.g., TTC). Let $\mathcal{A}_t = \{A_{t,1}, A_{t,2}, \dots, A_{t,k}\}$ be this partition of the agent set $A_t$, such that $A_t = \bigcup_{i=1}^k A_{t,i}$ and $A_{t,i} \cap A_{t,j} = \emptyset$ for $i \neq j$. Each $A_{t,i}$ represents the subset of agents belonging to group $i$ at time $t$. Then we get a new scene partition $\mathcal{G}_t = \{G_{t,1}, G_{t,2}, \dots, G_{t,k}\}$.

During training with IDS, we randomly sample a subset of these groups' indices. Let $\mathcal{I} \subseteq \{1, \dots, k\}$ be the indices of sampled groups, drawn according to a predefined sampling distribution $P_{\text{sample}}(\mathcal{I})$. The set of agents included in this sampled training instance at time $\tau$ is $A_{ \tau, \mathcal{I}} = \bigcup_{i \in \mathcal{I}} A_{\tau,i}$, where $A_{\tau,i}$ denotes the agents of group $i$ at time $\tau$. We then construct the sequence of sampled scene instances $G_{t-T_{hist}+1:t+T_{pred}}(\mathcal{I}) = \{ (S_{\tau}(\mathcal{I}); M^{\text{GT}}, L_{\tau}^{\text{GT}}) \mid \tau \in [t-T_{hist}+1, t+T_{pred}] \}$. Note that each $G_{\tau}(\mathcal{I})$ includes the full $M^{\text{GT}}$ and $L_{\tau}^{\text{GT}}$, but agent states ${S_{\tau}(\mathcal{I})}$ only for $A_{ \tau, \mathcal{I}}$. 

The neural network model $F_{\theta}$ is trained to predict the conditional probability distribution of the sampled future scene instances $\hat{G}_{t+1:t+T_{pred}}(\mathcal{I})$ given the sampled input scene history $G^{\text{GT}}_{t-T_{hist}+1:t}(\mathcal{I})$:
\begin{equation}
\begin{split}
P_{\text{model}}(\hat{G}_{t+1:t+T_{pred}}(\mathcal{I}) \mid G^{\text{GT}}_{t-T_{hist}+1:t}(\mathcal{I}); \theta) \\
\approx P_{\text{data}}(G^{\text{GT}}_{t+1:t+T_{pred}}(\mathcal{I}) \mid G^{\text{GT}}_{t-T_{hist}+1:t}(\mathcal{I}))
\end{split}
\end{equation}

The training objective under IDS is to minimize the expected negative log-likelihood of the true sampled future scene instances $G^{\text{GT}}_{t+1:t+T_{pred}}(\mathcal{I})$, averaged over the real-world data distribution $\mathcal{D}_{\text{data}}$ and the group sampling distribution $P_{\text{sample}}(\mathcal{I})$:
\begin{equation}
\begin{split}
\mathcal{L}(\theta) &=  - \mathbb{E}_{\hat{G} \sim \mathcal{D}_{\text{data}}} \mathbb{E}_{\mathcal{I} \sim P_{\text{sample}}(\mathcal{I})} \\&\left[ \log P_{\text{model}}(\hat{G}_{t+1:t+T_{pred}}(\mathcal{I}) \mid \right. \left. G^{\text{GT}}_{t-T_{hist}+1:t}(\mathcal{I}); \theta) \right]
\end{split}
\end{equation}
where $G^{\text{GT}}$ denotes the sequence of true scene instances $G_{t-T_{hist}+1:t+T_{pred}}^{\text{GT}}$.

\subsubsection{Marginal-to-Joint Simulation}
\label{sec:method:ids:simulation} 

During the closed-loop simulation phase, the model $F_{\theta}$ is presented with an arbitrary scene instance $G_t$ and its history $G_{t-T_{hist}+1:t}$. The task is to predict the full future scene instance $G_{t+1}$, corresponding to evaluating the conditional probability distribution $P_{\text{model}}(G_{t+1} \mid G_{t-T_{hist}+1:t}; \theta)$.

The power of IDS training lies in enabling Marginal-to-Joint simulation. Instead of learning from entire scenes, IDS trains the model on agent \emph{subsets}, forcing it to acquire a diverse vocabulary of \emph{conditional interaction primitives} ($\mathcal{P}$).  Let $\mathcal{P} = \{p_1, p_2, \dots, p_L\}$ denote this set, where each $p_i$ captures a specific local interaction pattern (e.g., yielding, following) learned from subset-based training. Conceptually, $\mathcal{P}$ spans the spectrum of interaction types observed across all training instances and their sampled subsets.

For an arbitrary scene $G_t$, the model predicts the next full state $G_{t+1}$ by first implicitly recognizing the combination of learned interaction primitives ($\mathcal{P}$) present in the current configuration and then synthesizing their future states. 
The model's ability to predict the \emph{joint} future state stems from its robust understanding of the \emph{marginal} dynamics of these smaller interaction groups ($p_i \in \mathcal{P}$).
By mastering these fundamental building blocks, the model can coherently predict complex scene dynamics, even for interaction combinations not explicitly seen during training. This approach transforms the challenge of synthesizing novel, complex interaction patterns within the same environment into a more tractable problem of combining learned primitives, thereby enhancing simulation robustness and long-term stability compared to methods that learn the high-dimensional joint distribution directly.

\subsection{Network Architecture}
The backbone of our IntersectioNDE is a multi-input, Transformer-based network designed for scene-level traffic simulation shown in Fig~\ref{fig:network}, employing an Encoder-Interaction-Prediction structure.

Diverse inputs are first processed by modality-specific encoders. These inputs include:
\begin{itemize}
    \item Historical Agent Trajectories: Represented as sequences of states over $T_{hist}$ history frames. For each agent $i$, its trajectory history is $H_i = \{h_{i,\tau} \mid \tau \in [t-T_{hist}+1, t]\}$, where $h_{i,\tau} \in \mathbb{R}^6$ includes 2D position $(\mu_x, \mu_y)$, 2D velocity $(v_x, v_y)$, and heading $\theta$ encoded as  $(\cos(\theta), \sin(\theta))$. The full historical agent trajectory input for the scene at time $t$ is $H_{t-T_{hist}+1:t} \in \mathbb{R}^{N \times T_{hist} \times 6}$.
    \item Agent Static Attributes: Including 3D size and one-hot encoded agent category. The full agent static attribute input is $A_{s} \in \mathbb{R}^{N \times 6}$.
    \item Routes: Representing planned or relevant routes for agents, typically as polylines. For $N_R$ route polylines, this input is $M_r \in \mathbb{R}^{N_R \times D_R}$, where $D_R$ is the feature dimension for each polyline representation.
    \item Traffic Lights: For $N_L$ traffic lights, the input is $M_d \in \mathbb{R}^{T_{hist} \times  N_L \times 3}$, where each light's state is one-hot encoded.
\end{itemize}

Modality-specific Encoders process these inputs to yield latent feature representations. The historical agent trajectories and static attributes are encoded into agent features $F_{agent} \in \mathbb{R}^{N \times D}$, where $D$ is the latent feature dimension. Static routes are encoded along with dynamic traffic light states into scene context features $F_{context} \in \mathbb{R}^{N_M \times D}$, where $N_M$ is the number of scene context elements.

Utilizing a Dual-Cross Attention Module, a central Interaction Network combines agent features with scene context features to produce enhanced agent features $\mathbb{R}^{N \times D}$ that are aware of their environment. These features are then processed by Transformer to model complex agent-agent dependencies across the scene, yielding refined agent features  $\mathbb{R}^{N \times D}$.

Finally, dedicated Prediction Heads for different agent categories process the refined agent features. For each agent category, a head predicts the parameters for the future kinematic state distribution over $T_{pred}$, outputting a tensor $\in \mathbb{R}^{N \times T_{pred} \times 6}$. This includes mean and standard deviation for position $(\mu_x, \mu_y, \sigma_x, \sigma_y)$ and $(\cos(\theta), \sin(\theta))$ for heading. An additional head predicts per-agent logits for high-level scene state transitions $\in \mathbb{R}^{N \times 3}$ (Stay/Leaving/Invalid).

\section{EXPERIMENT}
\subsection{Data Collection: City Crossings Dataset (CiCross)}

\begin{figure}[htbp]
    \centering
    \begin{subfigure}[b]{0.48\columnwidth}
        \centering
        \includegraphics[width=5cm, 
        trim=5cm 0cm 2cm 0cm, clip]{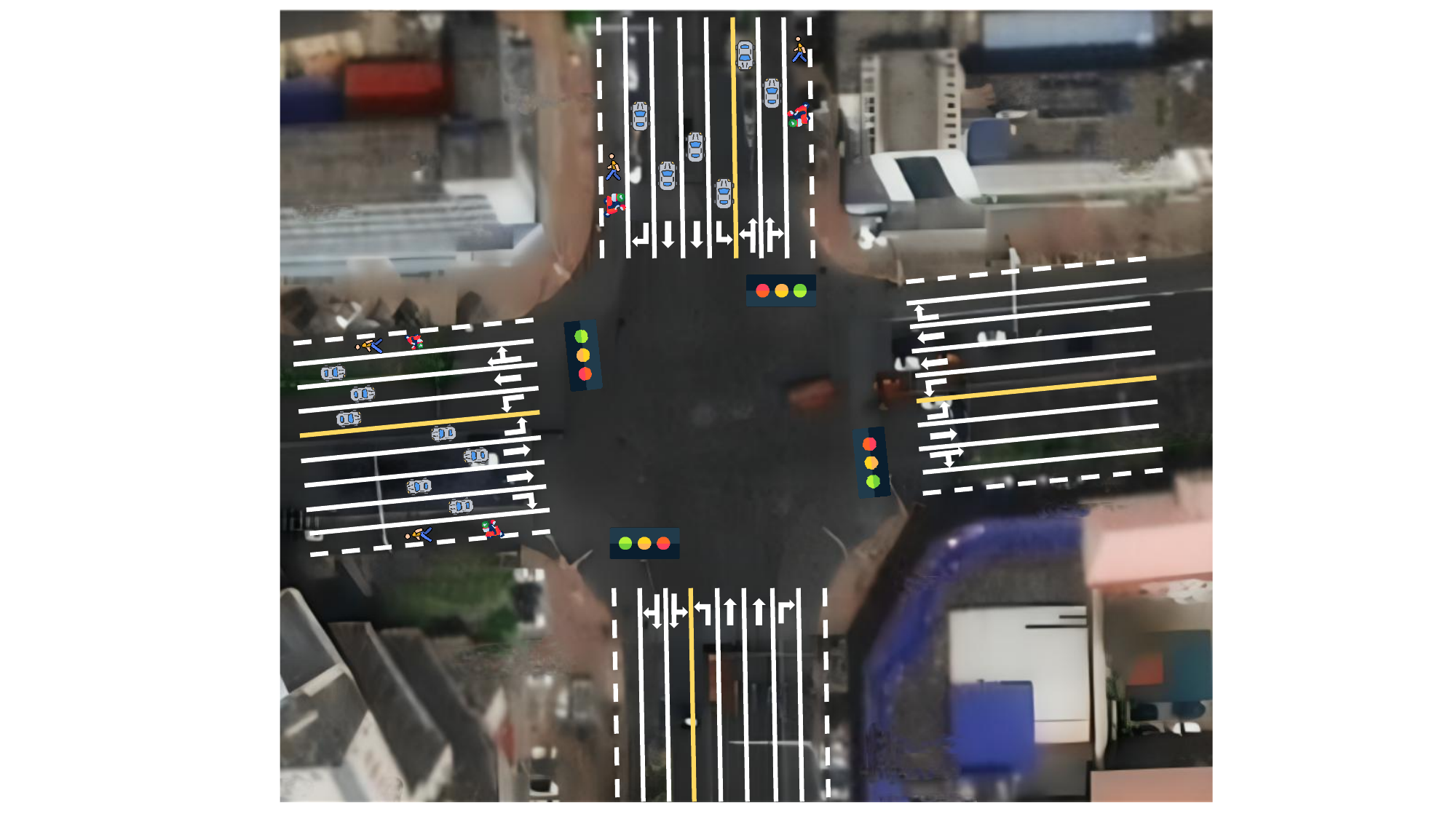}
        \caption*{a. CiCross Dataset Map}
    \end{subfigure}
    \hfill
    \begin{subfigure}[b]{0.48\columnwidth}
        \centering
        \includegraphics[width=\linewidth, trim=0cm 7.4cm 0cm 6cm, clip]{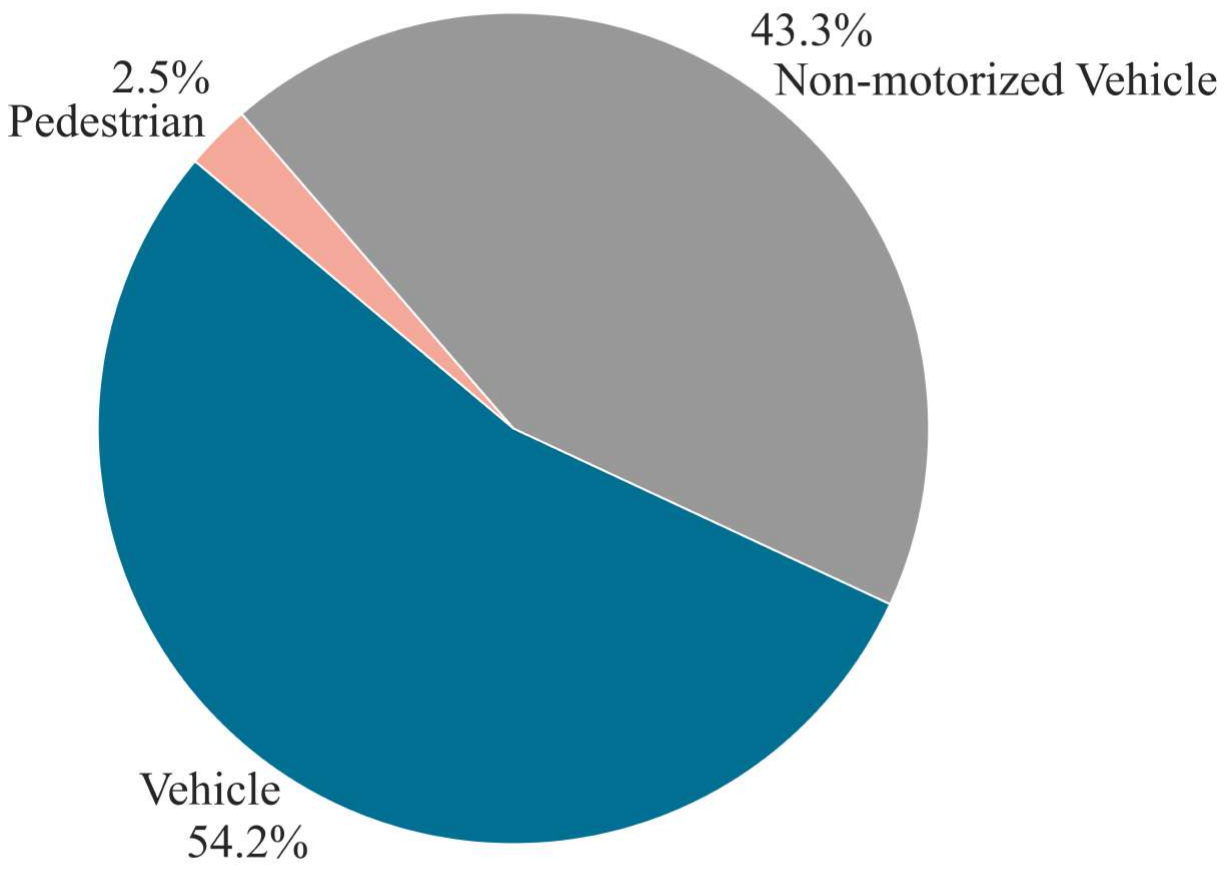}
        \caption*{b. Participant Type Ratio}
    \end{subfigure}
    \vspace{0.5em}
    \begin{subfigure}[b]{0.48\columnwidth}
        \centering
        \includegraphics[width=4.4cm,
        trim=0cm 6cm 0cm 5cm, clip]{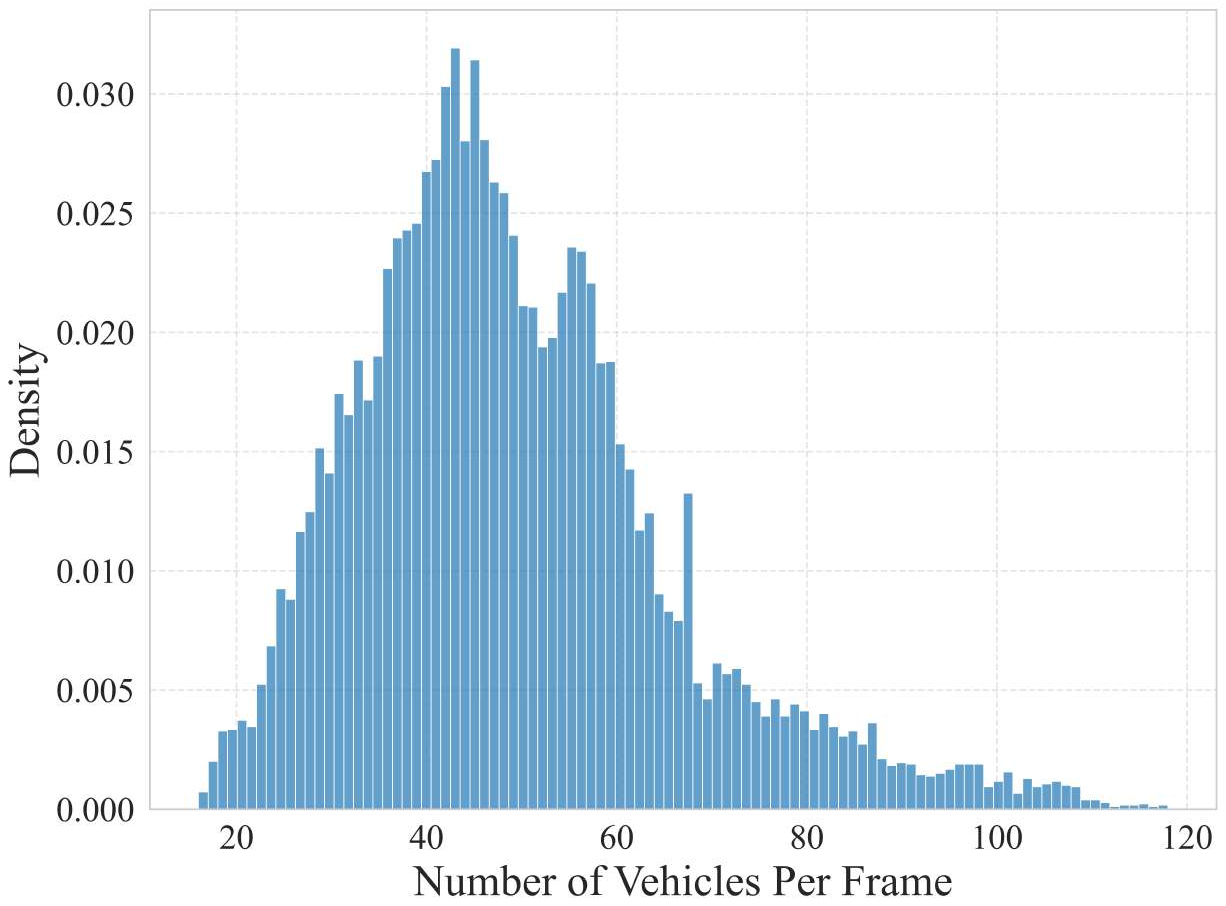}
        \caption*{c. Vehicle Count Distribution}
    \end{subfigure}
    \hfill
    \begin{subfigure}[b]{0.48\columnwidth}
        \centering
        \includegraphics[width=4.4cm,
        trim=0cm 6cm 0cm 5cm, clip]{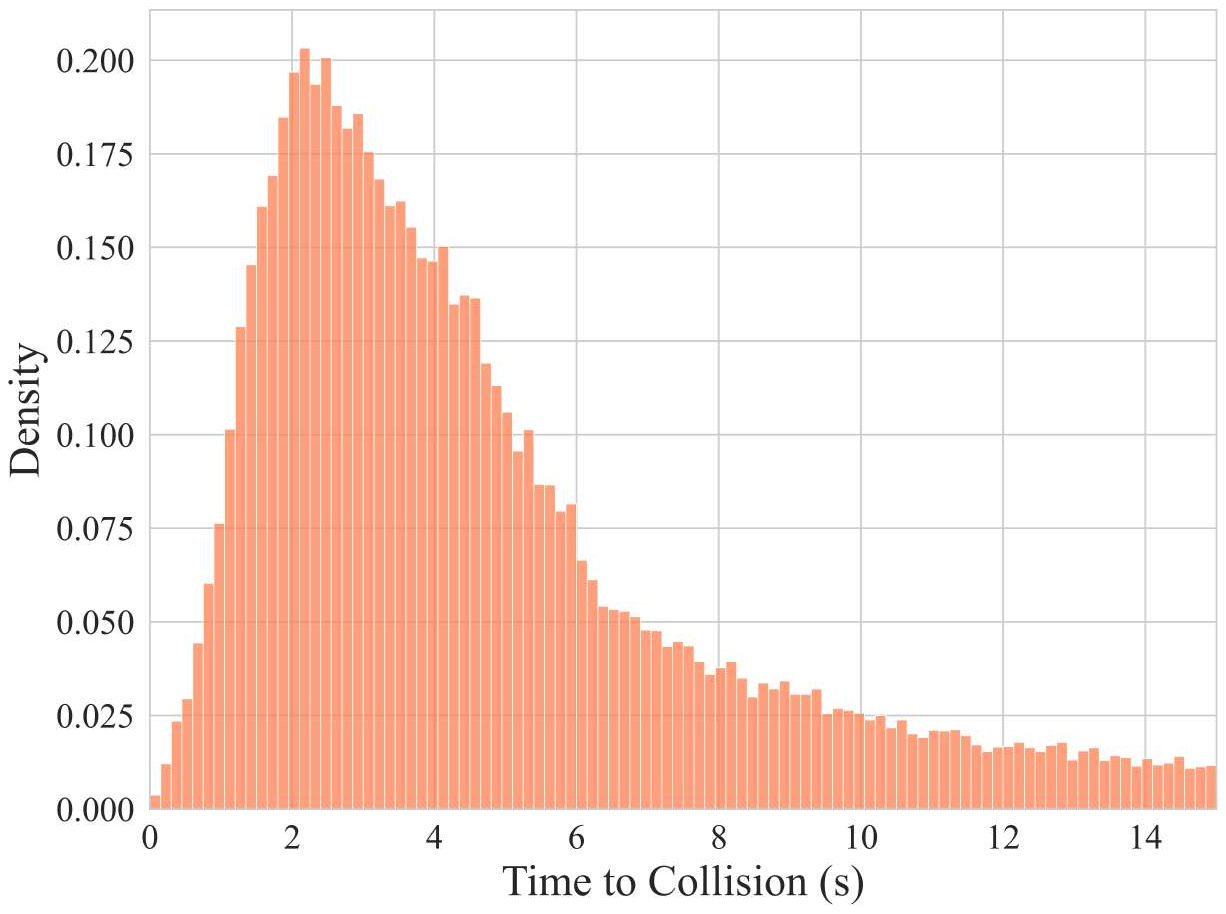}
        \caption*{d. TTC Distribution}
    \end{subfigure}

    \caption{Overview of CiCross map and key traffic statistics.}
    \label{fig:dataset}
\end{figure}

Our research utilizes City Crossings Dataset (CiCross), a high-fidelity dataset collected from a signalized urban intersection in a city in China, specifically curated to capture the challenging dynamics of dense, heterogeneous mixed traffic.
CiCross constitutes a large-scale collection, comprising approximately 700 hours of recorded traffic data from this single intersection. 
An illustration of the intersection layout is shown in Fig.~\ref{fig:dataset}(a), which includes four traffic lights and bidirectional multi-lane roads.
For this study, we use a 23.6-hour subset, which includes 212,344 frames at 2.5 Hz and 56,578 unique agent instances.

Over the entire subset, we tracked 56,578 agent instances. The distribution across categories (Motorized Vehicles - Category 1, Non-Motorized Vehicles (NMVs) - Category 2, Pedestrians - Category 3) is shown in Fig~\ref{fig:dataset}(b), with 54.2\% Motorized Vehicles, 43.3\% NMVs, and 2.5\% Pedestrians. 
The substantial proportion and sheer number of NMVs and pedestrians, combined with high agent density per frame (Fig~\ref{fig:dataset}(c)), significantly distinguishes CiCross even this subset from mainstream datasets and highlights its relevance for challenging urban mixed traffic scenarios. 
Furthermore, the Time-to-Collision (TTC) distribution, shown in Fig.~\ref{fig:dataset}(d), peaks around 2 seconds, highlighting the frequent and high-risk nature of the recorded scenarios.

\subsection{Experimental Setup}
Experiments were conducted on a subset of the CiCross dataset, comprising 212,344 frames (approximately 23.6 hours) recorded at 2.5 Hz from a complex urban intersection. The data was chronologically split into training (148,641 frames), validation (42,468 frames), and testing (21,235 frames) sets. Agent trajectories were normalized relative to a local scene origin for numerical stability. IntersectioNDE was trained with IDS (using TTC as the interaction standard) on a single NVIDIA RTX 4090 GPU using a history length $T_{\text{hist}} = 10$ and a prediction horizon $T_{\text{pred}} = 10$. Data augmentation included translation, rotation, shift, and trajectory error injection. Closed-loop rollouts were performed autoregressively with a 1-frame step size, sampling the next state from the predicted distribution. New agents were initialized from real test data, while existing agents were updated using model predictions.

\subsection{Closed-loop Simulation Evaluation}
\begin{figure}[htbp]
    \centering
    \begin{subfigure}[b]{0.32\columnwidth}
        \centering
        \includegraphics[width=\linewidth]{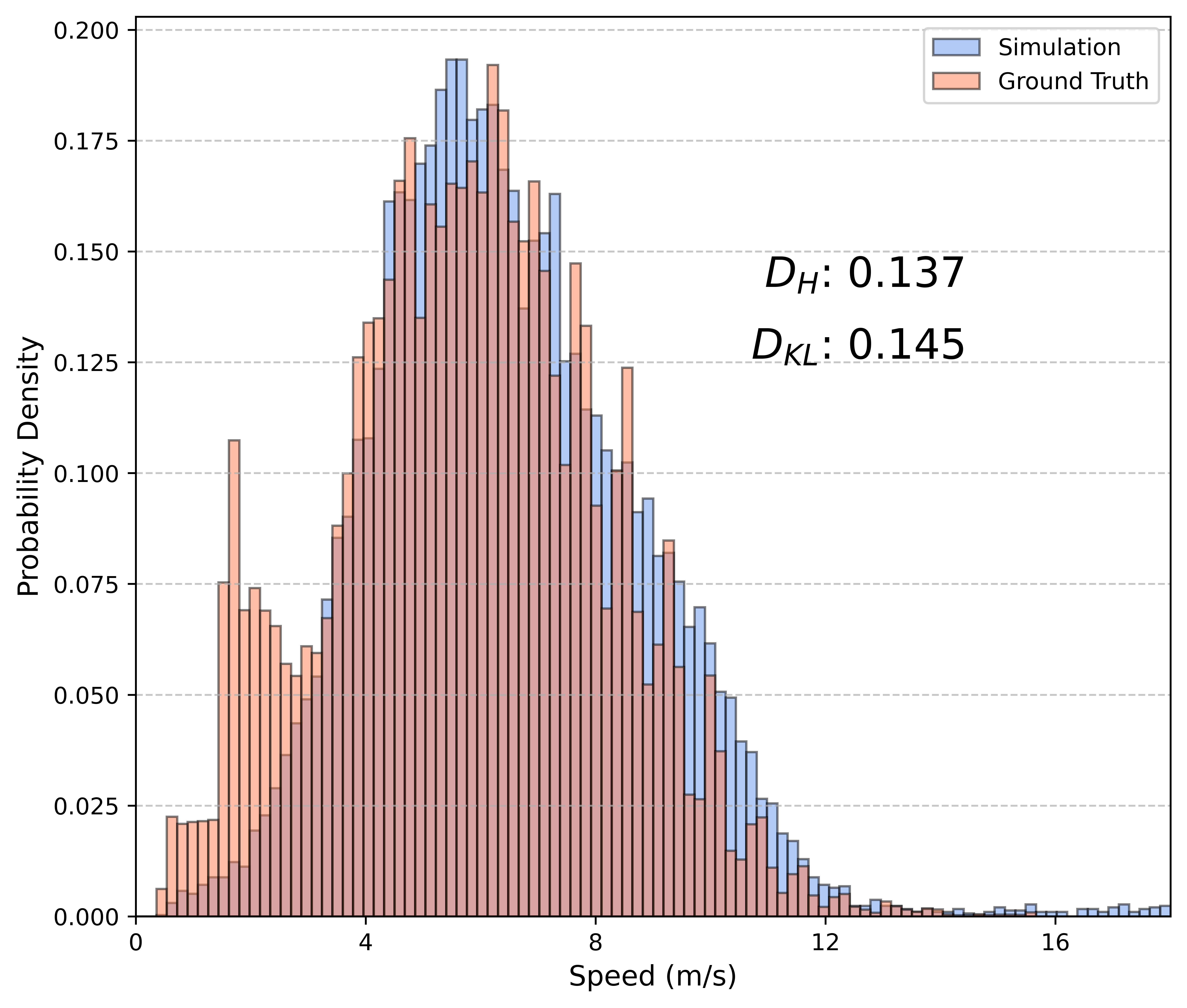}
        \caption{Motorized Vehicles}
        \label{fig:speed_vehicle}
    \end{subfigure}
    \hfill
    \begin{subfigure}[b]{0.32\columnwidth}
        \centering
        \includegraphics[width=\linewidth]{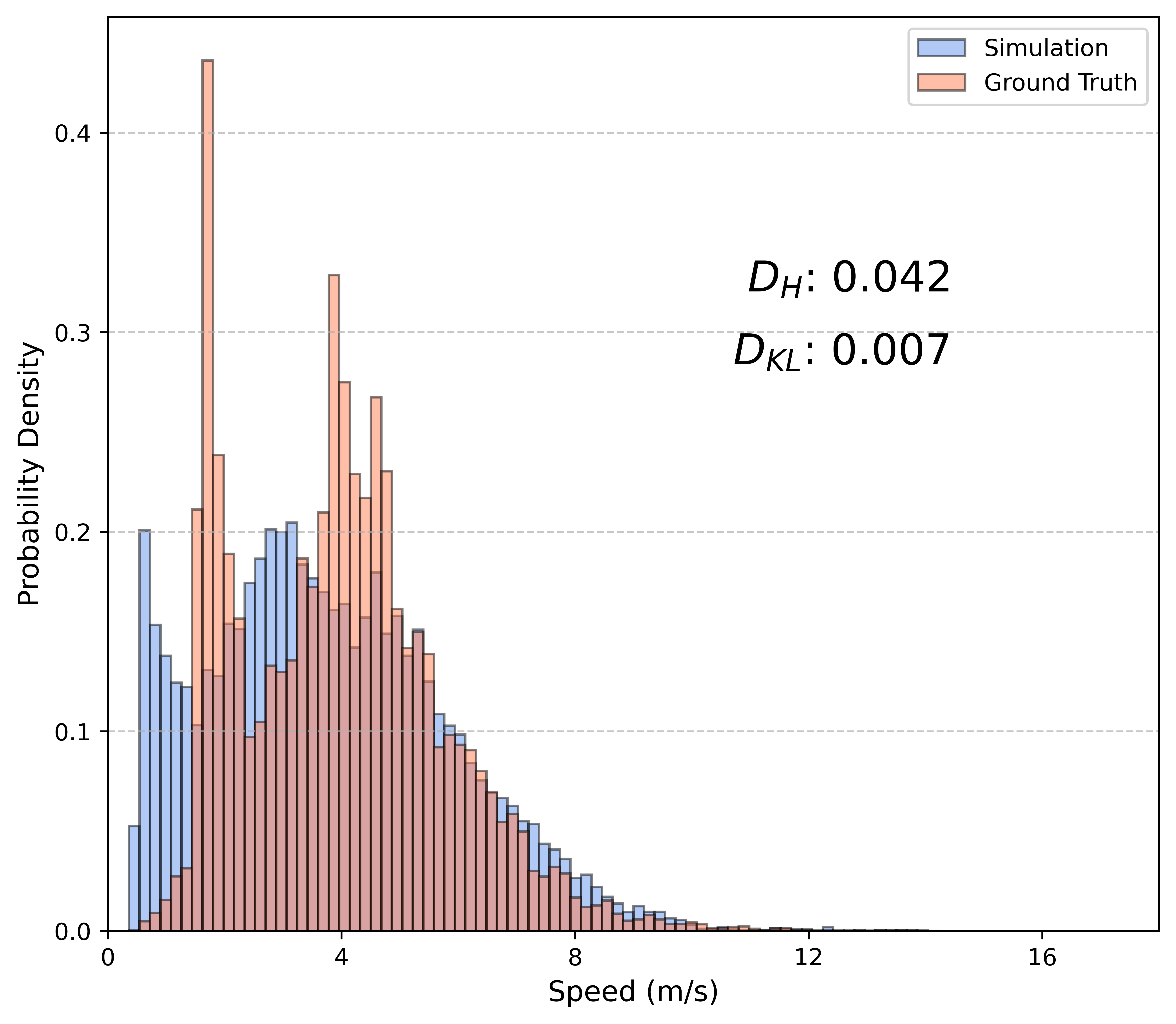}
        \caption{NMVs}
        \label{fig:speed_nmv}
    \end{subfigure}
    \hfill
    \begin{subfigure}[b]{0.32\columnwidth}
        \centering
        \includegraphics[width=\linewidth]{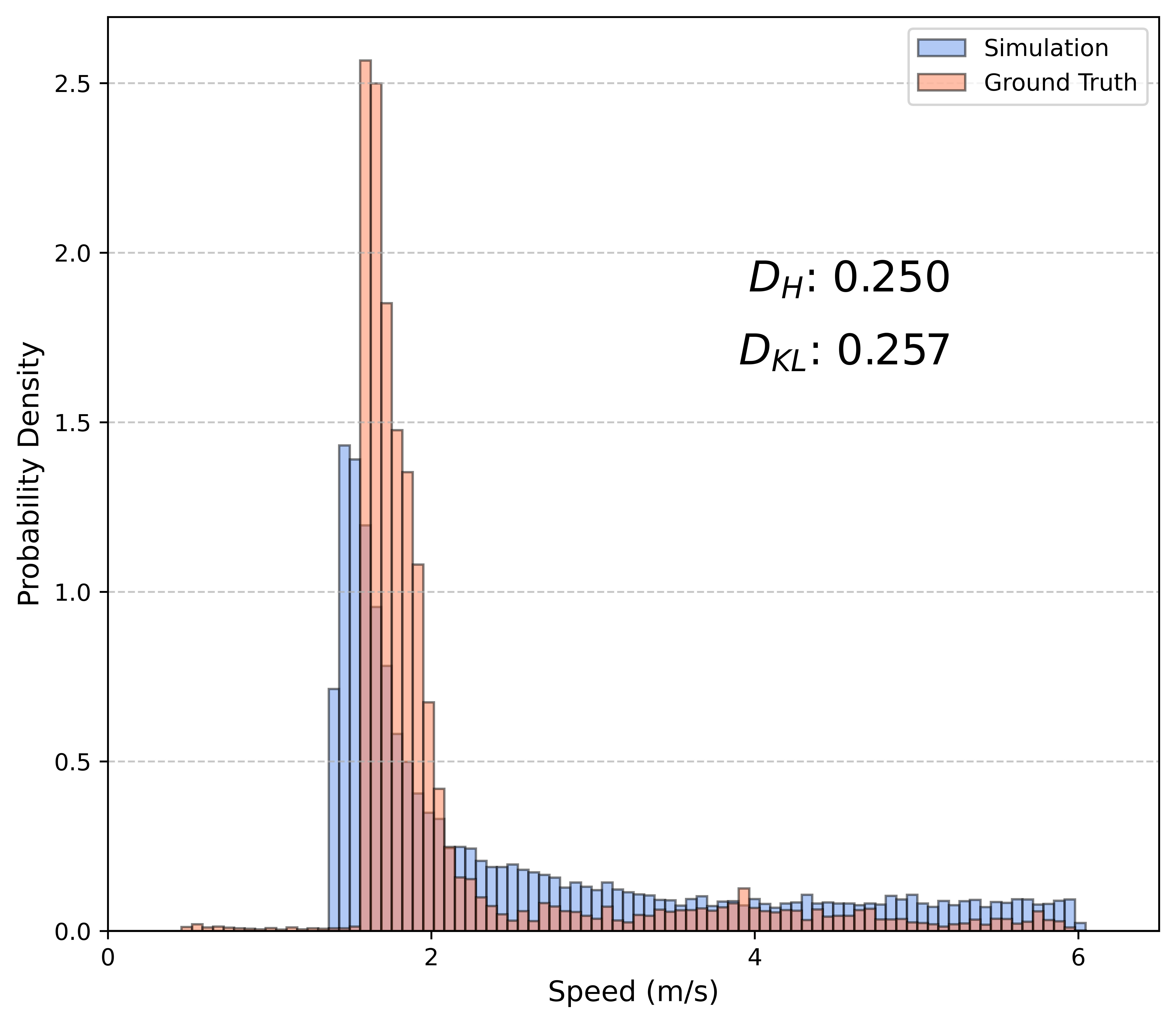}
        \caption{Pedestrians}
        \label{fig:speed_pedestrian}
    \end{subfigure}
    \caption{Speed Distribution of Different Participants}
    \label{fig:speed}
\end{figure}

\begin{figure}[htbp]
    \centering
    \begin{subfigure}[b]{0.32\columnwidth}
        \centering
        \includegraphics[width=\linewidth]{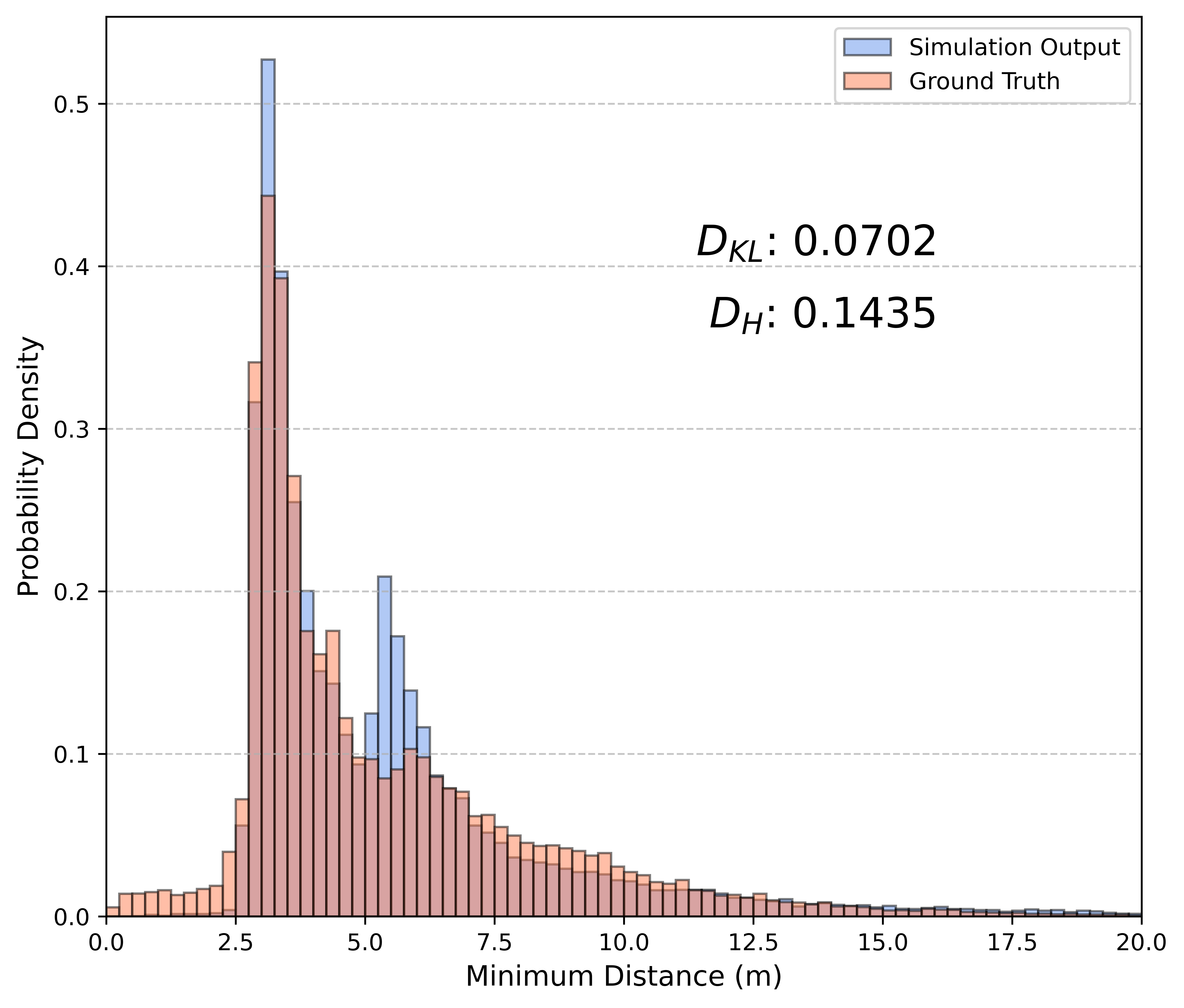}
        \caption{Motorized Vehicles}
        \label{fig:dist_vehicle}
    \end{subfigure}
    \hfill
    \begin{subfigure}[b]{0.32\columnwidth}
        \centering
        \includegraphics[width=\linewidth]{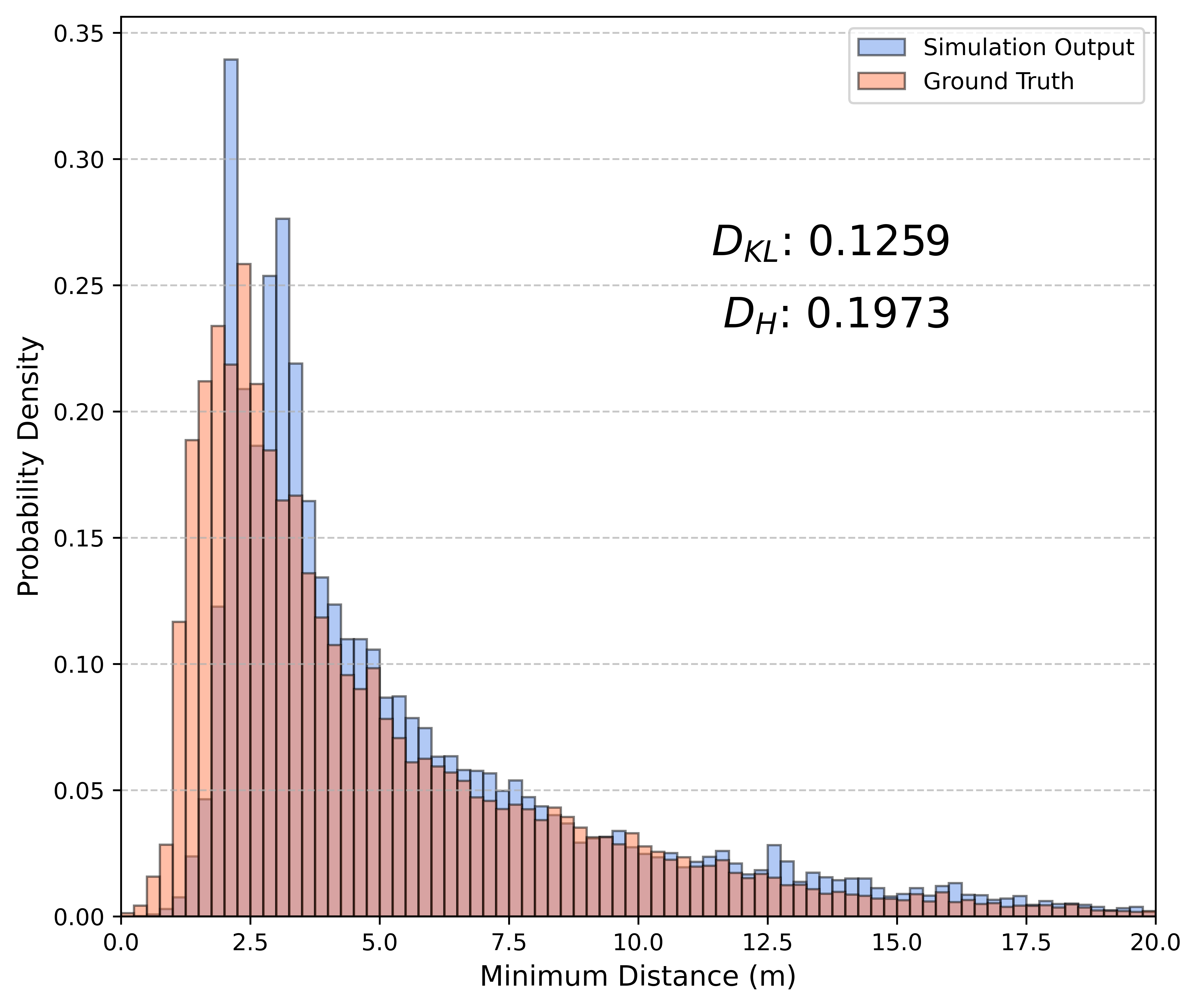}
        \caption{NMVs}
        \label{fig:dist_nmv}
    \end{subfigure}
    \hfill
    \begin{subfigure}[b]{0.32\columnwidth}
        \centering
        \includegraphics[width=\linewidth]{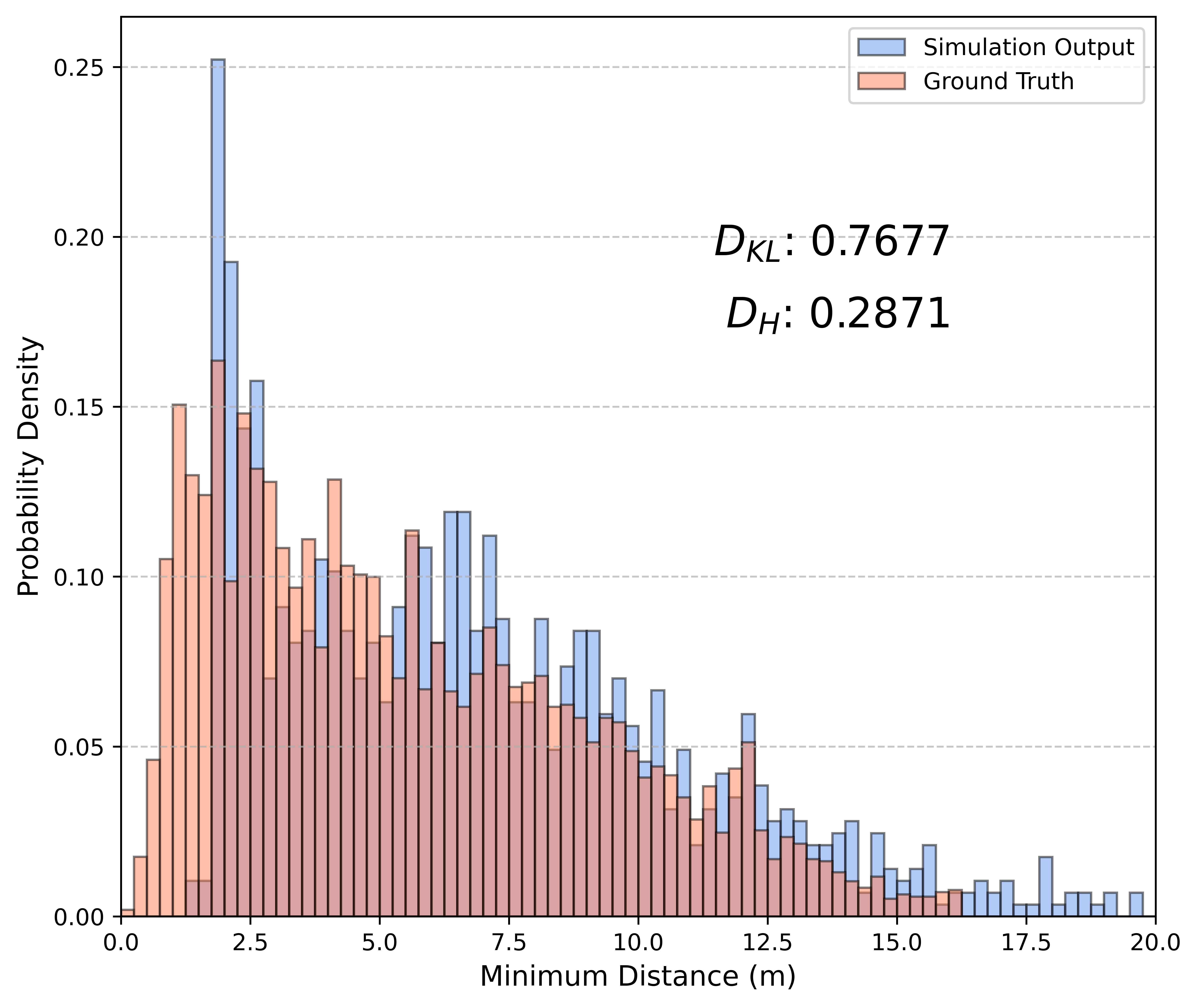}
        \caption{Pedestrians}
        \label{fig:dist_pedestrian}
    \end{subfigure}
    \caption{Closest Distance Distribution of Different Participants}
    \label{fig:min_distance}
\end{figure}

We evaluate simulation fidelity by comparing the distributions between the simulated and real data. Specifically, we analyze the distributions of agent speeds and closest distances, with corresponding distribution plots shown in Fig~\ref{fig:speed} and \ref{fig:min_distance}:
\begin{itemize}
\item Velocity: Distributions of agent speeds.
\item Closest Distance: Distributions of minimum distances between pairs of interacting agents (e.g., within a certain radius).
\end{itemize}

\begin{table}[htbp!]
\centering
\caption{Ablation Study Results}
\label{tab:ablation_study_with_avg}
\scriptsize
\begin{adjustbox}{width=\columnwidth}
\begin{tabular}{c c c c c}
\toprule
\textbf{Method} & \textbf{Participant} & \textbf{ADE↓} & \textbf{FDE↓} & \textbf{Missing Rate↓} \\
\midrule
\multirow{4}{*}{\makecell[c]{Dual-Cross Attention\\Without IDS}}
    & Motorized Vehicle    & 0.9047 & 1.6526 & 0.2086 \\
    & NMV         & 1.2864 & 2.4415 & 0.4553 \\
    & Pedestrian  & 1.2197 & 2.0536 & 0.3732 \\
    & \textbf{Average}   & 1.0131     & 1.8463     & 0.2699 \\
\midrule
\multirow{4}{*}{\makecell[c]{Dual-Cross Attention \\With IDS (TTC=0s)}}
    & Motorized Vehicle     & 0.6702 & \textbf{1.2466} & \textbf{0.1742} \\
    & NMV         & 1.0155 & 2.0130 & 0.3452 \\
    & Pedestrian  & 1.0224 & 1.6627 & 0.2342 \\
    & \textbf{Average}   & 0.7680 & 1.4324 & 0.2151 \\
\midrule
\multirow{4}{*}{\makecell[c]{Dual-Cross Attention\\With IDS (TTC=1s)}}
    & Motorized Vehicle     & \textbf{0.6693} & 1.2496 & 0.1750 \\
    & NMV         & \textbf{0.9869} & 1.9694 & \textbf{0.3310} \\
    & Pedestrian  & \textbf{1.0086} & \textbf{1.6150} & 0.2386 \\
    & \textbf{Average}   & \textbf{0.7590} & \textbf{1.4216} & \textbf{0.2120} \\
\midrule
\multirow{4}{*}{\makecell[c]{Dual-Cross Attention\\With IDS (TTC=2s)}}
    & Motorized Vehicle     & 0.7407 & 1.4113 & 0.1842\\
    & NMV         & 1.0780 & 2.1299 & 0.3698 \\
    & Pedestrian  & 0.9470 & 1.6165 & 0.2267 \\
    & \textbf{Average}   & 0.8362 & 1.5861 & 0.2292 \\
\midrule
\multirow{4}{*}{\makecell[c]{Dual-Cross Attention\\With IDS (TTC=4s)}}
    & Motorized Vehicle     & 0.7689 & 1.4692 & 0.1857 \\
    & NMV         & 1.0611 & 2.0838 & 0.3589 \\
    & Pedestrian  & 1.0915 & 1.8302 & 0.2645 \\
    & \textbf{Average}   & 0.8518 & 1.6179 & 0.2272 \\
\midrule
\multirow{4}{*}{\makecell[c]{Uni-Cross-Attention\\With IDS (TTC=1s)}}
    & Motorized Vehicle     & 0.6955 & 1.3159 & 0.1777 \\
    & NMV         & 0.9996 & \textbf{2.0016} & 0.3369 \\
    & Pedestrian  & 1.0266 & 1.6316 & \textbf{0.2144} \\
    & \textbf{Average}   & 0.7835 & 1.4834 & 0.2160 \\
\bottomrule
\end{tabular}
\end{adjustbox}
\end{table}

\subsection{Ablation Study}

To validate our contributions, we first establish a strong baseline. 
Our current work builds upon the insights from prior research, such as NeuralNDE~\cite{yan2023learning}, but targets a different scenario: a large-scale, mixed-traffic urban intersection. 
Thus, our baseline (``Without IDS'' in Table~\ref{tab:ablation_study_with_avg}) is a powerful scene-aware Transformer framework that we tailored specifically to handle the high agent density and model heterogeneous interactions inherent to this challenging setting.
The primary ablation then isolates the additional gains from our core methodological innovation --- the Interaction Decoupling Strategy --- when applied to this purpose-built backbone.

As shown in Table~\ref{tab:ablation_study_with_avg}, all IDS-based models outperform this baseline, confirming the strategy's overall effectiveness. 
We further analyzed the sensitivity of IDS to its TTC threshold, a critical hyperparameter. We tested thresholds of 0s, 1s, 2s, and 4s, a range informed by our dataset's TTC distribution (Fig~\ref{fig:dataset}(d)). A 0s threshold proves overly restrictive (\textit{interaction sparsity}), while higher thresholds (2s, 4s) introduce noise (\textit{interaction redundancy}). The 1s threshold strikes an effective balance, best capturing salient interactions.

Building on the optimal configuration, we also confirmed our Dual-Cross Attention outperforms a simpler Uni-Cross variant. Finally, the substantial improvement in collapse time (Table~\ref{tab:collapse_time}) demonstrates the crucial role of our full model in ensuring long-term simulation robustness.

\begin{table}[htbp]
\centering
\caption{Average Collapse Time Comparison}
\label{tab:collapse_time}
\scriptsize  
\begin{tabular}{l c}
\toprule
\textbf{Method} & \textbf{Average Collapse Time (s)} \\
\midrule 
Dual-Cross Attention(TTC=1s) & 895 \\
Without IDS   & 15 \\
\bottomrule
\end{tabular}
\end{table}

\begin{figure}[!hbp]
    \centering
    \includegraphics[width=\columnwidth,
    trim=0cm 3.5cm 0cm 1.5cm, clip]{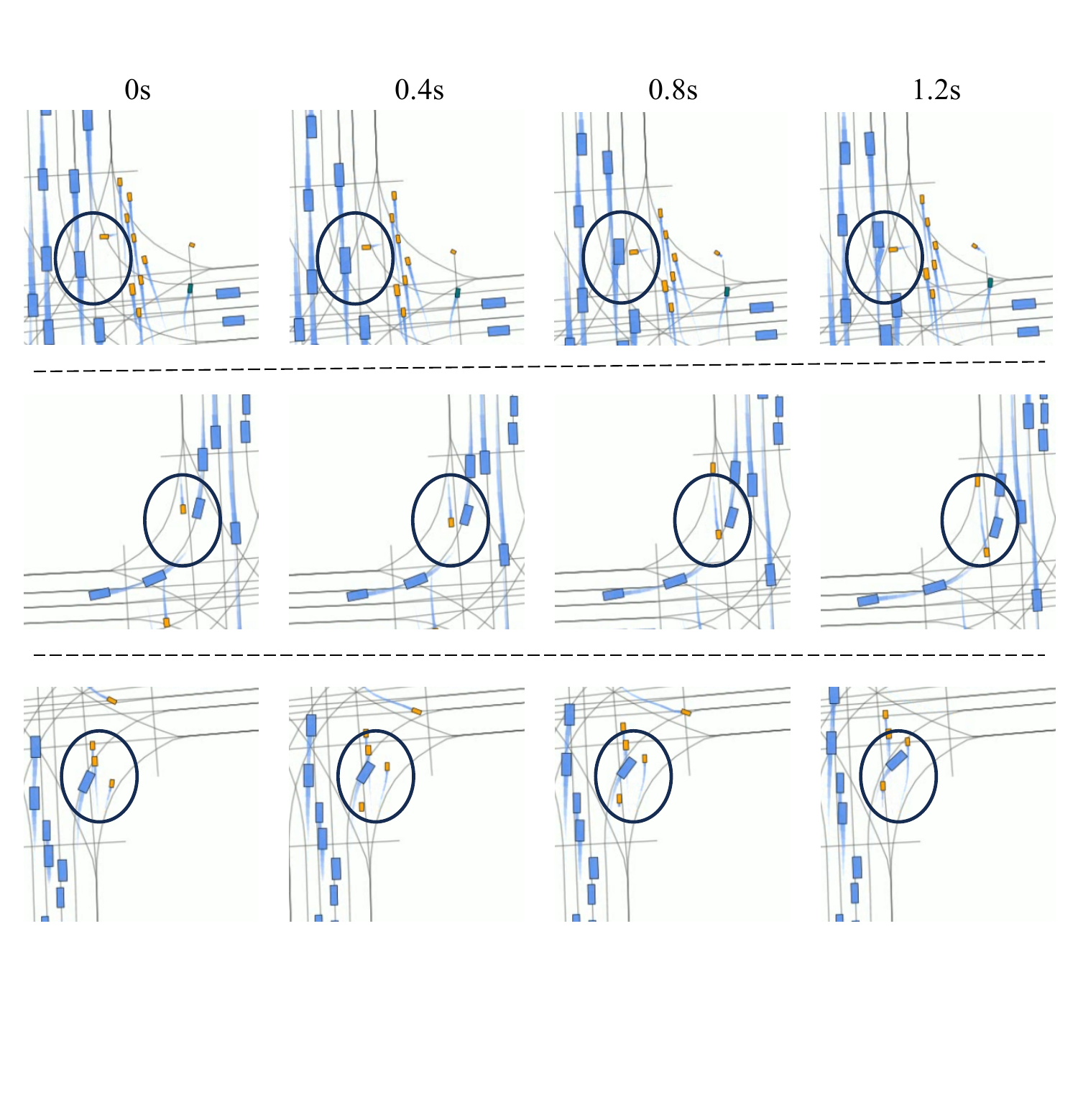}
    \caption{Simulation Case Study.}
    \label{fig:case_study}
\end{figure}

\subsection{Case Study}
To provide qualitative insights into IntersectioNDE's capabilities in simulating complex urban traffic, we present detailed case studies of representative scenarios from the CiCross test set.
We selected three typical interaction scenarios for analysis, as shown in Fig~\ref{fig:case_study}.

In Fig~\ref{fig:case_study}, blue denotes motorized vehicles and yellow denotes NMVs. 
In the first scenario, a non-motorized vehicle is running a red light and slows down or pauses due to obstruction from motorized vehicles approaching perpendicularly. 
In the second scenario, the motorized vehicle decelerates to yield. 
In the third scenario, the motorized vehicle encounters a stream of NMVs while making a right turn and responds by quickly turning through maneuvers such as yielding and accelerating.

\section{CONCLUSION}
In this work, we introduce CiCross, a novel dataset for heterogeneous urban interactions, and IntersectioNDE, a data-driven scene-level simulator. Leveraging marginal-to-joint simulation via our Interaction Decoupling Strategy (IDS), IntersectioNDE improves generalization to unseen interaction combinations within the same scene and achieves significantly enhanced simulation robustness and long-term stability in capturing complex urban traffic distributions.

\section*{Acknowledgment}
This research is supported by the National Key R\&D Program of China (2023YFB2504400), the Beijing Natural Science Foundation under Grant QY24260, and the Beijing Natural Science Foundation under Grant L243025.

\bibliographystyle{IEEEtran}
\bibliography{references}
\end{document}